\documentclass[journal]{IEEEtran}
\ifCLASSINFOpdf
\else
\fi
\usepackage{tabularx}
\usepackage{makecell}
\usepackage{graphicx}
\usepackage{subfig}
\usepackage{caption}
\usepackage[font=footnotesize]{caption}
\usepackage{amsmath}
\usepackage[switch,pagewise]{lineno} 
\usepackage{lipsum} 

\begin{document}
%
\title{Predicting Driver Takeover Time in Conditionally Automated Driving}
%
%
%

\author{{Jackie Ayoub, Na Du, X. Jessie Yang, Feng Zhou}
\thanks{Jackie Ayoub and Na Du contributed equally. Corresponding author: Feng Zhou.}
\thanks{J. Ayoub and F. Zhou are with the Department of Industrial and Manufacturing, Systems Engineering, The University of Michigan, Dearborn, 4901 Evergreen Rd. Dearborn, MI 48128 USA (e-mail: \{jyayoub, fezhou\}@umich.edu).}
\thanks{N. Du is with the Department of Informatics and Networked Systems, School of Computing and Information, University of Pittsburgh, 135 N Bellefield Ave, Pittsburgh, PA 15213 USA (e-mail: na.du@pitt.edu).}
\thanks{X. J. Yang is with Industrial and Operations Engineering, University of Michigan, Ann Arbor, 500 S State St, Ann Arbor, MI 48109 USA (e-mail: xijyang@umich.edu).}
\thanks{Manuscript received Nov 17, 2020; revised xxx 26, 2020.}}

%
%

\markboth{IEEE Transactions on Intelligent Transportation Systems,~Vol.~xx, No.~xx, October~2020}%
{Ayoub \MakeLowercase{\textit{et al.}}: Predicting Takeover Time in Conditional Automated Driving}
%



\maketitle

\begin{abstract}
 It is extremely important to ensure a safe takeover transition in conditionally automated driving. One of the critical factors that quantifies the safe takeover transition is takeover time. Previous studies identified the effects of many factors on takeover time, such as takeover lead time, non-driving tasks, modalities of the takeover requests (TORs), and scenario urgency. However, there is a lack of research to predict takeover time by considering these factors all at the same time. Toward this end, we used eXtreme Gradient Boosting (XGBoost) to predict the takeover time using a dataset from a meta-analysis study \cite{zhang2019determinants}. In addition, we used SHAP (SHapley Additive exPlanation) to analyze and explain the effects of the predictors on takeover time. We identified seven most critical predictors that resulted in the best prediction performance. Their main effects and interaction effects on takeover time were examined. The results showed that the proposed approach provided both good performance and explainability. Our findings have implications on the design of in-vehicle monitoring and alert systems to facilitate the interaction between the drivers and the automated vehicle. 
\end{abstract}


%
\IEEEpeerreviewmaketitle

\section{Introduction}
%
%
%
%
\IEEEPARstart{T}he SAE Level 3 automation allows drivers to take hands off the wheels, move feet off the pedals, and engage in non-driving-tasks (NDTs) when the automated mode is on \cite{sae}. However, when the vehicle reaches its operational limit and is not able to handle the driving scenario, the driver is required to take over control of the vehicle within a given time to negotiate the scenario safely (see \cite{ayoub2019manual, zhou2020takeover, zhou2020driver, du2020psychophysiological}). Takeover time is one of the critical measures quantifying the performance of takeover transitions. 

Previous literature evaluated takeover timeliness and takeover quality to assess drivers' takeover performance \cite{mcdonald2019toward,du2020examining,gold2016taking}.
Takeover time assesses the temporal dimension of takeover performance and is typically defined as the time between the TOR and the first maneuver of takeover actions, (e.g., gaze response time, eyes-on-road time, hands-on-wheel response time) \cite{zhang2019determinants, gold2013take}. Different thresholds have been used to capture the first takeover maneuver and the most commonly used threshold is 2$^{\circ}$ of steering rotation or 10\% of pedal inputs, whichever is quick \cite{gold2016taking,Louw2015,zeeb2016take}. Previous studies have investigated the effects of different factors on takeover time \cite{du2020evaluating,zeeb2017steering,gold2016taking}. For example, drivers responded quickly with less cognitive workload \cite{du2020evaluating} and alcohol consumption increased takeover time \cite{wiedemann2018effect}.

While those relationships provide an estimation on the direction takeover time changes in different conditions, they did not show what the estimated takeover time was under different takeover scenarios by examining various variables at the same time. Without knowing that, the system may not be able to provide critical situation awareness information needed (e.g., why a TOR is issued and how the takeover should be handled) at the right time to help negotiate the takeover process \cite{zhou2021using}. Thus, it is critical to build a computational model to accurately predict takeover time during the transitions from automated driving to manual driving.    

In this paper, we developed an explainable machine learning model based on XGBoost (eXtreme Gradient Boosting) \cite{xgboost} and SHAP (SHapley Additive exPlanations) \cite{lundberg2017unified,lundberg2020local} to predict takeover time in conditionally automated driving by aggregating 129 previous studies. XGBoost is one of the most successful machine learning models based on gradient boosting machines and has been widely applied in different areas \cite{torlay2017machine,gumus2017crude}. Thus, we used XGBoost to predict takeover time with all the possible variables identified in the literature. However, XGBoost is a black-box machine learning model, which makes it difficult to understand the contributions of different critical factors in predicting takeover time. Therefore, in order to explain the XGBoost model, we used SHAP, which has proved to provide locally accurate and consistent explanations \cite{lundberg2020local} based on Shaply values in game theory \cite{shapley1953value}. After identifying the most important variables, we were able to select only a small number of critical variables to predict takeover time with better performance and explainability (globally and locally) compared to previous studies. Globally, the model was able to rank the involved variables based on their contributions to takeover time prediction. Locally, the model explained individual instances with good accuracy. Such insights can be used to develop an adaptive in-vehicle alert system during the takeover transition period to improve driving safety in conditionally automated driving. As a summary, the contributions of this study are: 
\begin{itemize}
  \item We modeled the relationships between numerous predictor variables and takeover time using explainable machine learning models (i.e., XGBoost and SHAP).
  \item We used XGBoost to improve the predictive power of the model compared to other models (e.g., linear regression).
  \item We used SHAP to improve the explainability of XGBoost by feature selection and by providing both global (i.e., main and interaction effects) and local explanations (i.e., individual instance explanations).
  \item We identified critical relationships between predictor variables and takeover time, which paves the way to personalize takeover time prediction in specific scenarios.
\end{itemize}

\section{Related Work}
\subsection{Factors Influencing Takeover Time}
Prior research has examined many factors that influence drivers’ takeover time, such as takeover lead time \cite{du2020evaluating,wan2018effects}, TOR modality \cite{petermeijer2017take,naujoks2014effect}, NDTs \cite{zeeb2017steering,wandtner2018effects}, driving environments \cite{gold2016taking,du2020evaluating}, and drivers' characteristics \cite{zeeb2017steering,clark2017age,harvey2006neo}.
Takeover lead time refers to the time to collision at the time of the TOR \cite{mcdonald2019toward}. According to previous meta-analyses \cite{mcdonald2019toward,gold2018modeling}, an increase of 1 s in takeover lead time generally led to a 0.2 to 0.3 s increase in takeover time. Yet, such effects may disappear when the lead time is short. It may be because the short lead time triggers reflexive and quick responses. With regard to TOR modality, multi-modal TORs led to shorter takeover time compared to uni-modal ones \cite{petermeijer2017take,naujoks2014effect}. Among uni-modal TORs, visual TORs engendered longer takeover time than auditory or vibrotactile ones \cite{petermeijer2017take}.

NDTs influence takeover time through different perspectives. Compared to NDTs involving handheld devices, NDTs with mounted devices led to shorter takeover time because drivers could directly control the vehicles without putting down the devices in their hands \cite{wan2018effects,wandtner2018effects,zhang2019transitions}. While visual NDTs lengthened takeover time, auditory NDTs produced similar takeover time compared to conditions with no NDTs \cite{wandtner2018effects,radlmayr2014traffic}.
When NDTs induced drivers' different types of emotions and cognitive load, their effects on takeover time became complex \cite{du2020examining,zeeb2016take,du2020evaluating,zeeb2017steering,bueno2016different}. For example, some studies found that drivers' cognitive load and emotional valence/arousal did not influence takeover time \cite{du2020examining,du2020evaluating,bueno2016different}, while others found that higher cognitive load lengthened the reaction time in steering maneuvers but not braking maneuvers \cite{zeeb2016take,zeeb2017steering}. Further exploration is needed to address how NDTs interact with other variables to influence takeover time. 

Driving environments such as traffic density, scenario types, and weather are other important variables that influence takeover time \cite{gold2016taking,du2020evaluating,radlmayr2014traffic,du2020scenario,melnicuk2021effect}. When the traffic was in the same direction, studies found that heavy traffic increased takeover time \cite{gold2016taking,radlmayr2014traffic}. However, heavy oncoming traffic was found to reduce eyes-on-road reaction time when drivers were in low cognitive load \cite{du2020evaluating}. With regard to scenario types, Du et al. \cite{du2020scenario} found that lane changing scenarios induced shorter takeover time compared to lane keeping scenarios because lane changing scenarios had more contextual cues for drivers to process and the situations looked  more urgent.

Drivers' characteristics, such as age and automation experience, also influence takeover time \cite{clark2017age, zeeb2015determines,korber2016influence,li2018investigation}. For example, Zeeb et al. \cite{zeeb2015determines} found that repetitive exposure to takeover transitions reduced drivers' hands-on-wheel and eyes-on-road reaction time. The effects of age on takeover time were not consistent. Li et al. \cite{li2018investigation} found that older adults had longer takeover time compared to younger adults. In contrast, Korber et al. \cite{korber2016influence} and Clark et al. \cite{clark2017age} did not found any significant differences of takeover time during different age groups. Further studies are needed to investigate the effects of age on takeover time.

Despite the fact that the studies reviewed provided useful information on how different variables influence takeover time, it is difficult to estimate the exact takeover time at a specific takeover scenario. 
Hence, it is necessary to develop computational models to predict drivers’ takeover time under various takeover conditions.

\subsection{Existing Models on Takeover Time Prediction}
Existing studies have utilized cognitive architecture, linear and non-linear regression to model driver takeover time \cite{zhang2019determinants,gold2018modeling,deng2019modeling}. Deng et al. \cite{deng2019modeling} built the queuing network-adaptive control of thought rational (QN-ACTR) cognitive architecture to model takeover time. The model used drivers' task-specific skills and knowledge as production rules and capacities of human mental processing as algorithms and parameters. The prediction reached $R^2$ of 0.96, root mean square error (RMSE) of 0.5 s, and mean absolute percentage error (MAPE) of 9\% in six empirical experiments. Gold et al. \cite{gold2018modeling} modeled takeover time using data from six driving simulator experiments (753 takeover situations). The thresholds of maneuver start were set as 2° steering wheel angle change or 10\% brake pedal actuation. According to variable analysis results, time-budget, traffic density, repetition, lane position and driver’s age were selects as predictor variables to model takeover time using a generalized non-linear model. The model obtained RMSE of 0.81 s, adjusted $R^2$ of 0.43, and deviation of  0.1 - 0.7 s using a validation dataset from another five additional studies. Zhang et al. \cite{zhang2019determinants} reviewed 129 studies (520 takeover events) on takeover time with SAE Level 2 - 3 automation. A linear mixed model was developed with automation level, TOR modality, NDT modality, hand occupation, situation urgency, and other road users as dummy variable inputs. The model showed goodness-of-fit measures (log likelihood) of 132.2, Akaike’s Information Criterion (AIC) of 136.2, and Schwarz’s Bayesian Criterion (BIC) of 144.6.

\section{Methodology}

\subsection{Dataset} 
The dataset used in this study was based on a meta-analysis study, investigating the variables affecting takeover time \cite{zhang2019determinants}. A total number of 129 studies were reviewed including conference proceedings, journal articles, technical reports, posters, and Ph.D. or master theses. 
\begin{figure}[tb!]
\centering
  {\includegraphics[width=0.7\linewidth]{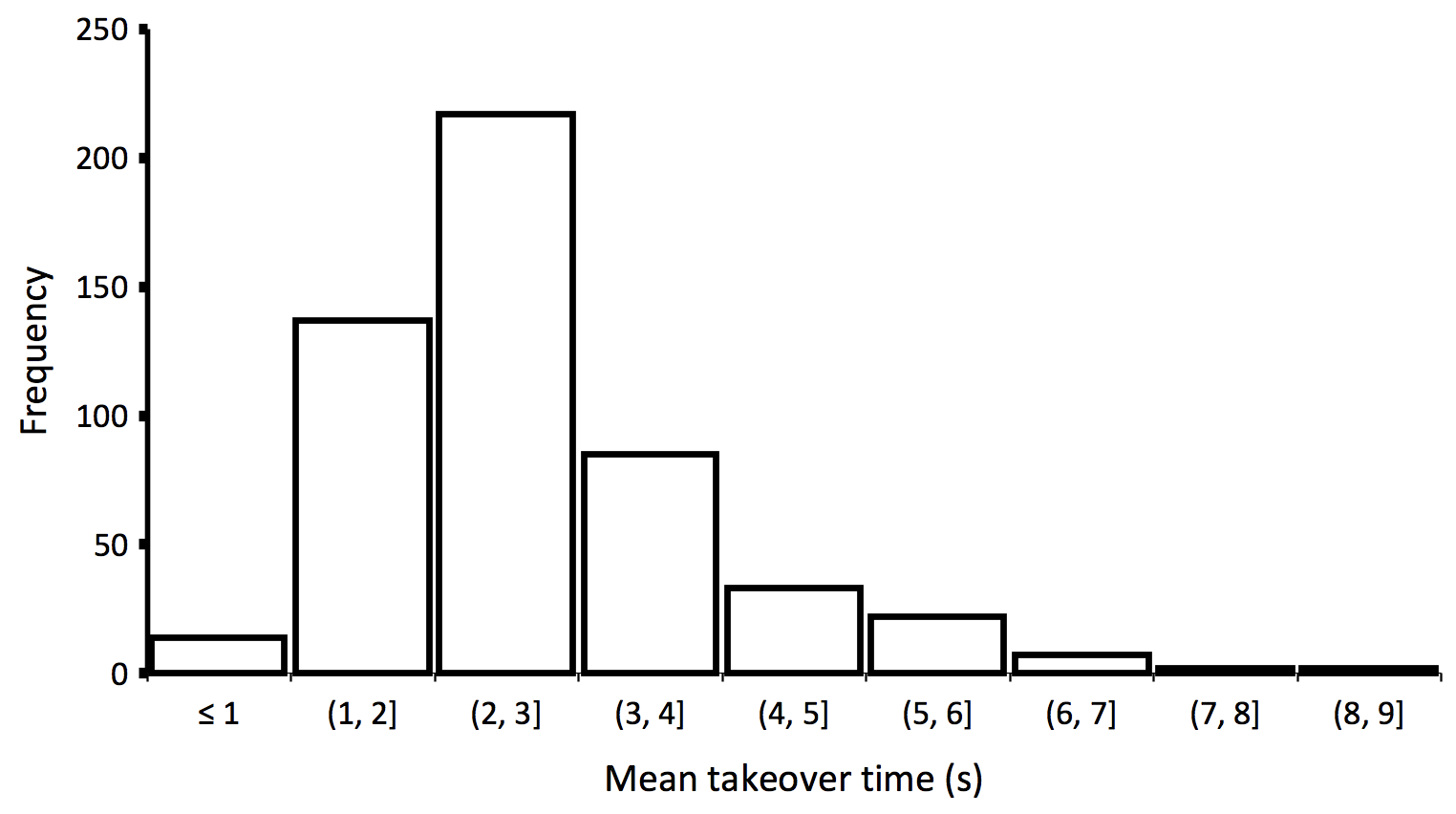}}\hfill
  \caption{The distribution of 519 takeover time  using a bin width of 1 s.}
  \label{fig:Ydistribution}
\end{figure}
As a result, 18 variables were identified to influence the takeover time as shown in Table \ref{table:variables}. The identified variables were related to drivers (i.e., age, cognitive load, and drivers' response), automated vehicles systems (i.e., level of automation, simulator fidelity), TOR modality (i.e., visual, auditory, and vibrotactile requests), NDTs (i.e., visual, auditory, and motoric NDTs), interactions with other road users, and the urgency of the situation (i.e., urgency, time budget to collision, and time budget to other boundaries). Only the studies that used the same calculation method of takeover time were selected. Takeover time was defined as the time interval between the initiation of the takeover stimulus (i.e., the onset of the TOR or the start of an environmental event that can initiate driver takeover) and the moment of driver intervention (by means of braking, steering, or button pressing).
The identified takeover time ranged from 0.69 s to 19.79 s and its distribution was shown in Fig. \ref{fig:Ydistribution}. Since only one takeover time value was larger than 8.7 s (i.e., 19.79 s), this value was considered as an outlier and was excluded from this study.

\begin{table*}[tb]
\centering
\caption{Variables identified through literature review that influenced takeover time.}
\label{table:variables}
\resizebox{17cm}{!}{%
\begin{tabular}{@{}lll@{}}
\hline \hline
\multicolumn{1}{c}{\textbf{Abbreviation}} & \multicolumn{1}{c}{\textbf{Feature Description}} & \multicolumn{1}{c}{\textbf{Scale}} \\ 
\hline 
AGE & Mean age of the participants. & Years \\
LAD & \begin{tabular}[c]{@{}l@{}}The level of automated driving where in L2, drivers’ are in charge of the monitoring task and in L3 and above,\\ drivers’ are not needed to monitor the  task.\end{tabular} & 0 = L2; 1 = L3 and above \\
SIM & \begin{tabular}[c]{@{}l@{}}The low fidelity simulator is either desktop based or computer monitors. The  medium fidelity simulator is either \\ a simulator with more than 120 deg  horizontal field of view or an instrumented-cabin simulator. \\The high fidelity  simulator is either a real car or a simulator with motion platform.\end{tabular} & \begin{tabular}[c]{@{}l@{}}0 = low fidelity; 1 = medium fidelity;\\  2 = high fidelity\end{tabular} \\
TOR\_V & Whether a visual TOR exists. & 0 = no; 1 = yes \\
TOR\_A & Whether an auditory TOR exists. & 0 = no; 1 = yes \\
TOR\_VT & Whether a vibrotactile TOR exists. & 0 = no; 1 = yes \\
TOR\_P & Whether a TOR is present. & 0 = no; 1 = yes \\
NDT\_V & Whether the NDT is visual. & 0 = no; 1 = yes \\
NDT\_A & Whether the NDT is auditory. & 0 = no; 1 = yes \\
NDT\_M & Whether a hand operation is needed to perform the NDT. & 0 = no; 1 = yes \\
NDT\_C & \begin{tabular}[c]{@{}l@{}}NDTs requiring working memory are considered as high cognitive load, or they are normal cognitive load.\end{tabular} & \begin{tabular}[c]{@{}l@{}}0 = normal cognitive load; 1 = \\ high cognitive load\end{tabular} \\
HAND & Whether a device is handheld while performing the NDT. & 0 = hands-free; 1 = handheld \\
NDT\_P & Whether a non-driving task is present. & 0 = no; 1 = yes \\
TBTC & \begin{tabular}[c]{@{}l@{}}Time budget from the initiation of the TOR until the collision with  an obstacle.\end{tabular} & Seconds \\
TBTB & \begin{tabular}[c]{@{}l@{}}Time budget from the start of the TOR until reaching the boundaries of the automated system \\other than collision.\end{tabular} & Seconds \\
URG & Whether a potential urgency exists. & \begin{tabular}[c]{@{}l@{}}0 = low urgency (TBTC $>$ 15 s); \\1  = medium urgency (8 s $<$ TBTC $\leq$ 15 s);\\ 2 = high urgency (TBTC $\leq$ 8 s)\end{tabular} \\
DRE & \begin{tabular}[c]{@{}l@{}}Drivers’ response in a low complexity situation include taking over control on a straight road by stabilizing \\the vehicle in its lane. In a medium complexity situation, the TOR requires a specific driver response \\(e.g., road narrowing, curvy road, road construction, decelerating vehicle ahead). In a high \\ complex situation, drivers’ need to decide whether to brake or steer in response to the event.\end{tabular} & \begin{tabular}[c]{@{}l@{}}1 = low complexity; 2 = medium\\  complexity; 3 = high complexity\end{tabular} \\
IRU & \begin{tabular}[c]{@{}l@{}}Whether an interaction with other road users exists during the takeover  process.\end{tabular} & 0 = no; 1 = yes \\ 
\hline \hline
\end{tabular}
}
\end{table*}

Note the obtained takeover time was the mean takeover time reported in each condition of the reviewed studies, and it is the response variable to be predicted in our study using the XGBoost model. The ground truth is the mean takeover time collected in the meta-analysis study \cite{zhang2019determinants}. The corresponding independent variables associated with that takeover time were considered as the predictor variables. The prediction process was done by training the XGBoost model on predicting the ground truth of takeover time using the corresponding independent variables. The dataset was in a tabular format and the total number of instances for training and testing was 519. In order to reduce the percentage of the missing values, dummy variables were used for many variables. For instance, a visual TOR was coded as "0" if the TOR was not visual and "1" otherwise. Moreover, the time budget to collision and the time budget to other boundaries variables were combined into one variable since they both represented the time from the initiation of the TOR to reaching the longitudinal or lateral boundaries. Finally, missing values accounted for 9.67\%.

\subsection{XGBoost Model Development}
In this study, we used XGBoost \cite{xgboost} as a regression model to predict the takeover time. One of the motivations of using XGBoost in predicting the takeover time is that it provides a great performance with less computational resources. XGBoost has the advantage to model automatically high order interaction and to handle missing values in the dataset. Another motivation is that the predictions of the XGBoost model can be explained using SHAP. The combination of XGBoost and SHAP provides both good performance and explainability of the takeover prediction.
The XGBoost algorithm is a scalable learning algorithm composed of an ensemble of decision trees. The learning process was designed to minimize the objective function by iteratively splitting each tree and giving more weights to the accurate ones. The main idea of XGBoost is to build the regression model by adding one tree at a time to minimize the objective function. The objective function is composed of a loss function calculated between the predicted and actual values and a regularization term to avoid over-fitting.  
XGBoost is scalable to different applications and is able to handle missing values without an imputation preprocessing in this study. In addition, XGBoost builds trees using parallel computing, and thus it runs much faster than the traditional gradient tree boosting models \cite{chen2019}. 

Given a dataset $ D = \{x_i^k$, $y_i, i = 1, ..., N, k = 1, ..., M\}$, where $M$ is the number of the predictor variables (i.e., 18 in total in this study) and $N$ is the number of the observations (i.e., 519 observations in this study), the XGBoost algorithm aims to minimize the following objective function:
\begin{equation}
L^{(t)} = \sum_{i=1}^{n} l(y_i,\hat{y_i}^{(t-1)} + f_t(x_i^k)) + \Omega(f_t), \\
\end{equation}
where $n$ is the number of the training samples, $\hat{y_i}^t = \hat{y_i}^{(t-1)} + f_t(x_i^k)$ is the prediction at the $t$-th iteration, $f_t(x_i^k)$ is the $t$-th tree to be added, and $\Omega(f_t)$ is the regularization term to avoid over-fitting. 
The learning process starts by iterating over all the features of the training data. Then, a split reduction loss is calculated for each tree to obtain the optimal split having the minimum reduction loss. In order to make predictions, an output value is calculated for each leaf of the tree to make the final predictions. 

In this work, the XGBoost regressor was trained with a 10-fold cross-validation strategy and using an optimized search for hyperparameters. The optimized hyperparamters were: subsample, colsample, learning\_rate, max\_depth, n\_estimators. In addition, we ran the 10-fold cross-validation 100 times (i.e., random state $=0-99$) to stabilise and obtain optimal results from the XGBoost regressor. 

\subsection{XGBoost Model Evaluation}
To compare model performance, four metrics were used including RMSE, Mean Absolute Error (MAE), Adjusted $R^2$ (short for Adj. $R^2$), and correlation coefficient (Corr) between the predicted values and ground truth. RMSE is defined as the square root of the mean squares of the difference between the predicted and actual value, and it changes from $0$ to $\infty$ (see Eq. \ref{RMSE}). It indicates how close the observed data points are to the model's predicted value. The lower the RMSE the better the model fits the data. MAE is defined as the mean absolute difference between the predicted and actual values, and it changes from $0$ to $\infty$ (see Eq. \ref{MAE}). It indicates the average magnitude of the errors in a set of predictions, and the lower the MAE, the better the model predicts the mean takeover time. Adj. $R^2$ is a revised version of $R^2$ by adjusting the number of predictor variables. It measures the explanatory power of the regression model (see Eq. \ref{adjR2}). It usually changes from 0 to 1, with 0 indicating that the proposed model does not improve the prediction and 1 indicating perfect prediction. Corr represents the strength of the linear relationship between the predicted takeover time and the ground truth. It changes from -1 to +1, with -1 indicating a perfect negative correlation and +1 indicating a positive correlation.

\begin{equation}
RMSE = \sqrt{\frac {1}{N}  \sum_{i=1}^{N}(y_i - \hat{y_i})^2}, \\
\label{RMSE}
\end{equation}

\begin{equation}
MAE = \frac {1}{N}  \sum_{i=1}^{N}|y_i - \hat{y_i}|, \\
\label{MAE}
\end{equation}

\begin{equation}
Adj. R^2 = 1 - (1 - R^2) \frac{N-1}{N-M-1},\\
\label{adjR2}
\end{equation}

\begin{equation}
Corr = \frac{\sum_{i=1}^{N} (\hat{y_i}-\bar{\hat{y}})(y_i -\bar{y})} {\sqrt{\sum_{i=1}^{N} (\hat{y_i}-\bar{\hat{y}})^2 \sum_{i=1}^{N}(y_i -\bar{y})^2}}, \\
\label{Corr}
\end{equation}
where \textit{N} is the total number of the observations, \textit{M} is the total number of the predictor variables, $R^2 = \frac{\sum_{i=1}^{N} (\hat{y}_i-\bar{y})^2}{\sum_{i=1}^{N} (y_i - \bar{y})^2}$, $y_i$ represents the $i$-th actual value, $\mathit{\hat{y}_i}$ represents the $i$-th predicted value, $\mathit{\bar{y}}$ is the average of all the actual values, and $\bar{\hat{y}}$ is the average of the all the predicted values.

\subsection{SHAP Explanation}

Explanability is key to the adoption of a machine learning model. It improves people's trust in model predictions. Ayoub et al. \cite{Ayoub2021Modeling, ayoub2021combat} demonstrated that showing why and how the predictions were made was necessary for users to trust the model. SHAP explains model predictions by using the Shapley values from game theory \cite{shapley1953value}. The Shapley value represents the contribution of each variable in pushing the model prediction from the expected value (see Eq. \ref{phi}). The unit of the SHAP value is the same as the prediction space which is in seconds in this work.  
In the case of XGBoost, each predictor variable is assigned with an importance value based on their contribution to the final prediction. 

\begin{equation}
\phi_i(\upsilon) =  \sum _{S \subseteq N \backslash \{i\}} \frac{|S|!(M-|S|-1)!}{M !} (\upsilon(S \cup \{i\}) - \upsilon(S)),\\
\label{phi}
\end{equation}
where $N$ is the set of all the predictor variables, $\textit{M}$ is the total number of the predictor variables, \textit{S} is a subset of any coalition of the predictor variables and  $\mathit{\upsilon (S)}$ is the contribution of coalition \textit{S} in predicting takeover time in our study. 

To explain the predictions of XGboost, SHAP provides global and local explanations of model predictions. Global explanations provide information about variable importance ranking, main effects of predictor variables on the response variable, and interaction effects between predictor variables on the response variable, whereas local explanations help understand individual predictions by showing the variables contributing to the obtained prediction. The interaction effect is defined as the effect of the combined variables minus the main effects of the individual variables in Eq. \ref{ph}.

\begin{equation}
\begin{aligned}
&\phi_{i,j}(\upsilon) = \sum _{S \subseteq m \backslash \{i,j\}} \frac{|S|!(M-|S|-2)!}{M !} \\
& (\upsilon(S \cup \{i,j\}) - \upsilon(S \cup \{i\}) - \upsilon(S \cup \{j\}) + \upsilon(S)).\\
\label{ph}
\end{aligned}
\end{equation}

\section{RESULTS}
\subsection{XGBoost Performance}
The performance metrics of the XGBoost prediction model including RMSE, MAE, Adj. $R^2$ and Corr are shown in Table \ref{table:modelperformance} using 10-fold cross validation for 100 times with different random numbers (0-99) in Python 3.8. In addition, we took a data-driven method and used feature selection to identify the best performance of the model with the current dataset by adding one variable at a time following the importance ranking of the variables identified in Fig. \ref{fig:variableRank}. We continued this process until RMSE stopped decreasing. RMSE was used as our main performance measure since it is probably the most popular evaluation metric used in regression analysis. When the RMSE had the smallest value (with seven most important predictors in the model), all the other metrics, including MAE, Adj. $R^2$, and Corr had the best performance. Finally, we found that XGBoost performed the best when a combination of seven variables (i.e., URG, TBTC\&TBTB, AGE, HAND, TOR\_V, SIM, IRU was used.

\begin{table} [tb]
\centering
\caption{Mean value of the model performance measures while using different variables.}
\label{table:modelperformance}
\resizebox{9cm}{!} {
\begin{tabular}{l c c c c}
\hline\hline
\small
\textbf{Variables}  & \textbf{RMSE} $\downarrow$  & \textbf{MAE} $\downarrow$   & \textbf{Adj. $R^2$} $\uparrow$ & \textbf{Corr} $\uparrow$  \\ [0.2ex]
\hline
URG                                  & 1.119          & 0.830          & 0.187  & 0.717 \\
URG, TBTC\&TBTB                              & 0.965          & 0.709          & 0.394           & 0.815          \\
URG, TBTC\&TBTB, AGE                         & 0.851          & 0.560          & 0.528           & 0.867          \\
URG, TBTC\&TBTB, AGE\\HAND                   & 0.828          & 0.538          & 0.552           & 0.875          \\
URG, TBTC\&TBTB, AGE,\\HAND, TOR\_V           & 0.826          & 0.533          & 0.554           & 0.877          \\
URG, TBTC\&TBTB, AGE,\\HAND, TOR\_V,SIM      & 0.808          & 0.518          & 0.572           & 0.883          \\
URG, TBTC\&TBTB, AGE,\\HAND, TOR\_V,SIM, IRU & \textbf{0.806} & \textbf{0.505} & \textbf{0.573}           & \textbf{0.883}\\[1ex]
\hline \hline    
\end{tabular}}
\end{table}

\subsection{Global Explanation}
\subsubsection{Variable Importance}
\begin{figure}[tb!]
\centering
\includegraphics[width=.9\linewidth]{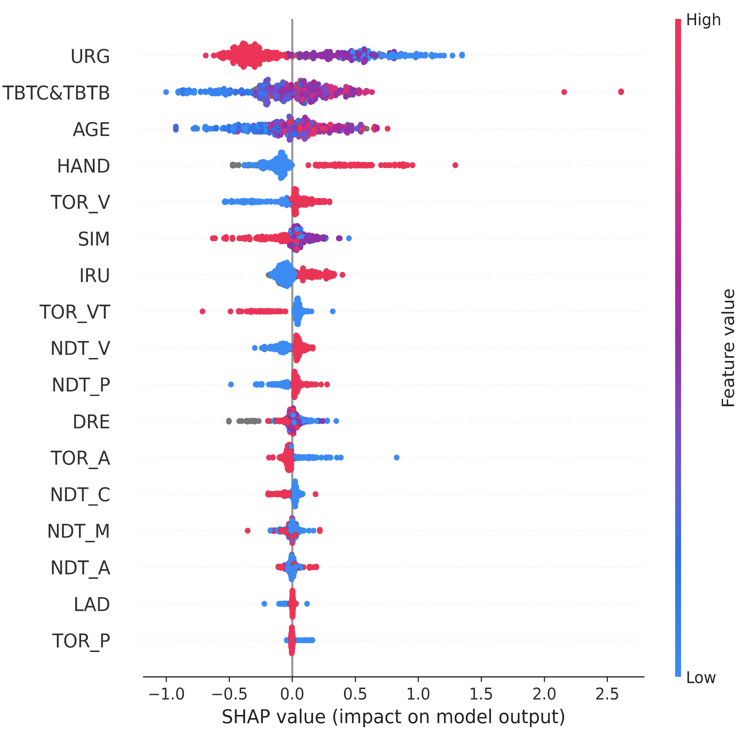}\hfill
\caption{SHAP summary plot.}
\label{fig:variableRank}
\end{figure}

To understand the importance of the variables in predicting takeover time, we examined the SHAP summary plot in Fig. \ref{fig:variableRank}. The SHAP values represent unique variable contribution to each prediction. The variables are ranked based on their global importance, i.e., $\sum_{i=1}^{N} |\phi_j ^{(i)}|$, where $\phi_j ^{(i)}$ represents the SHAP value of the $i$-th instance of the $j$-th variable. Each dot in the plot represents a SHAP value for its instance of one variable. It has four characteristics including (1) the color represents the value of that variable ranging from low (blue) to high (red), (2) the horizontal location represents the small or large effect of the variable on the prediction, (3) the vertical location represents the importance of a particular variable, and (4) the density shows the distribution of the variable in the dataset \cite{molnar2019}. In Fig. \ref{fig:variableRank}, we see that URG is the most important variable in takeover time prediction, followed by TBTC\&TBTB, AGE, and HAND, and so on. For example, a high level of urgency was shown to reduce the predicted takeover time by almost 1 s, whereas a low level of urgency increased the predicted takeover time by almost 1.5 s. Additionally, we saw variations in the tails of the variables, where the important variables had a wider tail compared to the less important ones with SHAP values centered around 0 (e.g., NDT\_M, NDT\_A, LAD, TOR\_P).         

\begin{figure}[!ht]
\centering
\includegraphics[width=1\linewidth]{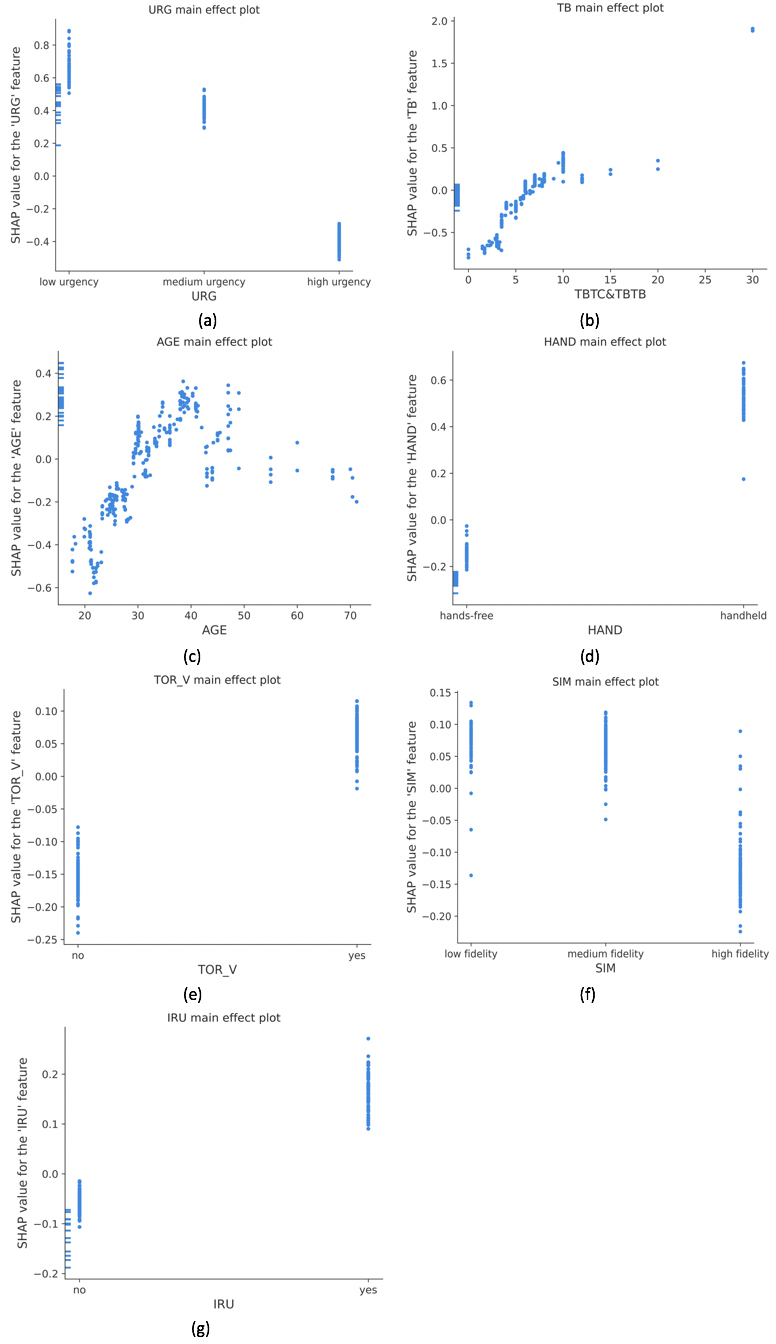}\hfill
\caption{SHAP main effect plots. (a) URG (b) TBTC\&TBTB (c) AGE (d) HAND (e) TOR\_V (f) SIM (g) IRU.}
\label{fig:Main Effect}
\end{figure}

\begin{figure}[!ht]
\centering
\includegraphics[width=1\linewidth]{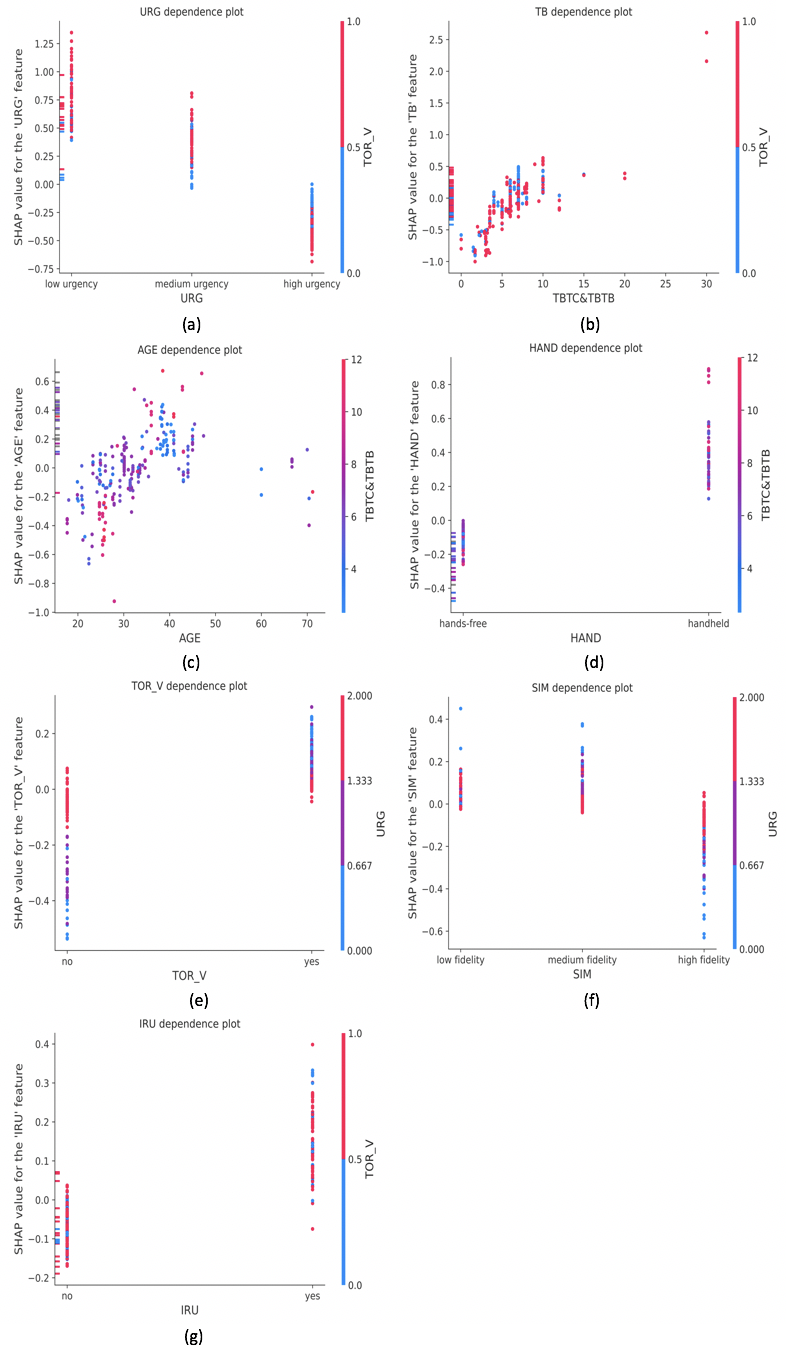}\hfill
\caption{SHAP dependence plots. (a) URG (b) TBTC\&TBTB (c) AGE (d) HAND (e) TOR\_V (f) SIM (g) IRU.}
\label{fig:Inter Effect}
\end{figure}

\subsubsection{Main and Interaction Effects}
To further understand the effects of the predictor variables on the takeover time, we explored the main and interaction effects of the most important variables (see Figs. \ref{fig:Main Effect} and \ref{fig:Inter Effect}). The vertical axis represents the SHAP value that shows the effects of the variables on the prediction and the horizontal axis represents the actual value of the variables. In the main effect plot for URG, the takeover time decreased with increased URG (see Fig.  \ref{fig:Main Effect}a). As for time budget, an increase in TBTC\&TBTB increased the takeover time (see Fig. \ref{fig:Main Effect}b). As for the variable AGE, an increasing takeover time was shown until AGE reached 45 years old where the takeover time started to decrease after it was larger than 45 years old (see Fig. \ref{fig:Main Effect}c). Hand holding a device while performing NDTs was shown to increase the takeover time (see Fig. \ref{fig:Main Effect}d). The existence of a visual TOR was shown to increase the takeover time (see Fig. \ref{fig:Main Effect}e). Driving simulators with a high fidelity seemed to reduce the takeover time compared to those with a low or medium fidelity as seen in Fig. \ref{fig:Main Effect}f. An interaction with road users increased the takeover time (see Fig. \ref{fig:Main Effect}g).

A certain extent of vertical dispersion at single values of the predictor variables was noticed in the main effect plots. Such dispersion represents the interaction effect of the predictor variables with other variables automatically identified by SHAP. To reveal these interaction effects, a second vertical column was added to the right side of the main effect plots (i.e., SHAP dependence plots) in Fig. \ref{fig:Inter Effect}. Each dot represents a single prediction from the dataset. The first vertical axis on the left represents the SHAP value of the studied variable, which represents how much this variable can change the output of the takeover time. The horizontal axis represents the value of the variable. The color corresponds to a second variable on the right that has an interaction effect with the variable. The interaction is presented as a distinct vertical pattern of coloring. By plotting the values of two predictor variables in the dataset, we can see how they interact. 
For instance, in Fig. \ref{fig:Inter Effect}a shows that among the situations with a low or medium level of urgency during the takeover, those with visual TORs increased the takeover time compared to those with non-visual TORs. However, among the takeover situations with a high level of urgency, those with non-visual TORs increased the takeover time compared to those with visual TORs. In Fig. \ref{fig:Inter Effect}b, among the situations with a low time budget to collision or to other boundaries (time budget $\leq$ 15 s or so), non-visual TORs increased the takeover time compared to the visual TOR, while no such effects were observed when the time budget $>$ 15 s or so.
As for the variable AGE, a larger time budget increased takeover time for those between 30 and 45 years old while a larger time budget decreased takeover time for those smaller than 30 years old as shown in Fig. \ref{fig:Inter Effect}c. 
As for the variable HAND, a larger time budget increased the takeover time when holding a device in their hands performing NDTs while a larger time budget tended to decrease the takeover time when the participants' hands were free during NDTs (see Fig. \ref{fig:Inter Effect}d). 
In addition, the interaction plot of TOR\_V showed that when the TORs were not visual, a high urgent situation increased the takeover time compared to a low urgent situation. However, when the TORs were visual, a high urgent situation decreased the takeover time compared to a low urgent situation (see Fig. \ref{fig:Inter Effect}e). 
As for the simulator fidelity, for takeover scenarios that took place in simulators with a low or medium fidelity, a high urgency decreased the takeover time while for those happened in simulators with a high fidelity, a high urgency increased the takeover time (see Fig. \ref{fig:Inter Effect}f). 
As for the variable IRU, there was not a clear interaction effect, possibly due to small effect size, though TOR\_V was selected (see Fig. \ref{fig:Inter Effect}g).

\subsection{Local Explanation}
\begin{figure}[tb!]
\centering
\includegraphics[width=1\linewidth]{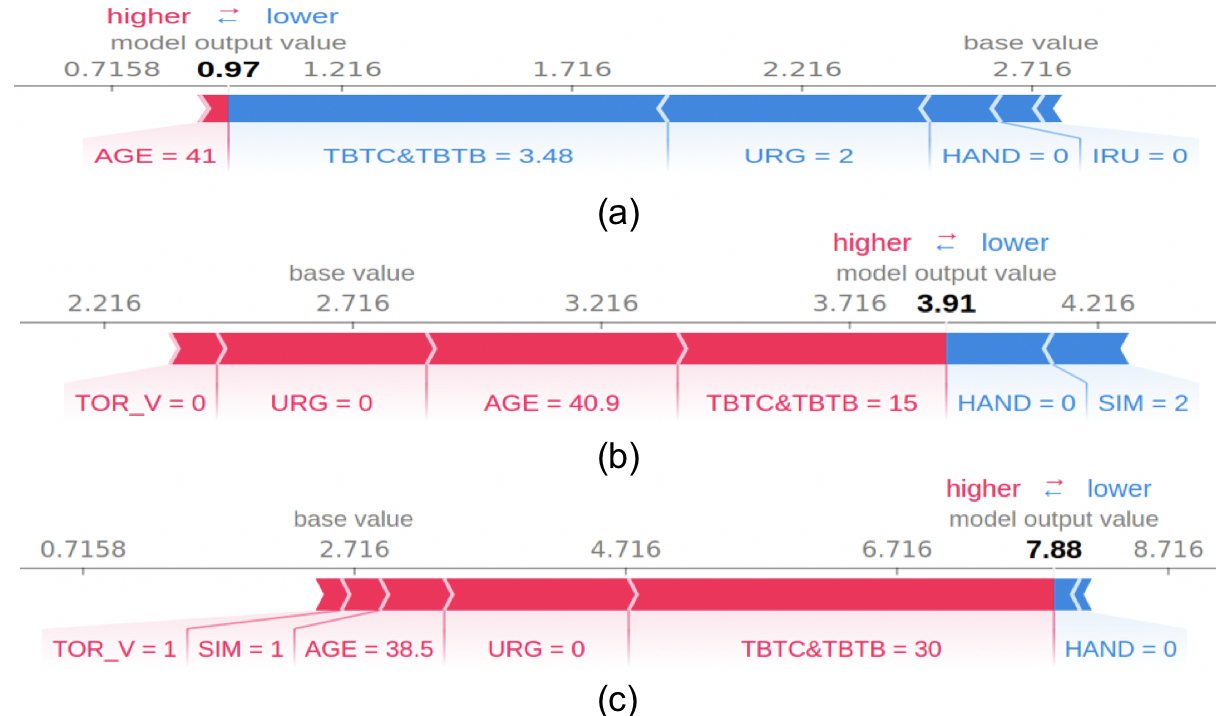}\hfill
\caption{SHAP individual explanations of takeover prediction for (a) ground truth = 0.89 s, (b) ground truth = 3.94 s, and (c) ground truth = 8.70 s.}
\label{fig:localMin}
\end{figure}

To explain individual predictions using SHAP, we randomly selected three cases as illustrated in Fig. \ref{fig:localMin}. The provided explanations show different variables pushing the predicted takeover time from the base value (i.e., average prediction over the training data) to the output value (i.e., predicted takeover time). Thus, each value of a specific predictor variable will either decrease or increase the predicted value from the base value (e.g., 2.716 in Fig. \ref{fig:localMin}a). For instance, the variables in red (e.g., TBTC\&TBTB=15 in Fig. \ref{fig:localMin}b) pushed the predicted takeover time higher than the baseline while those in blue (e.g., HAND = 0 in Fig. \ref{fig:localMin}c) pushed the predicted takeover time lower than the baseline. By aggregating all the contributions to predicting takeover time from the base value, the actual prediction for these three particular instances are obtained as shown in Fig. \ref{fig:localMin}. 

In the first example (see Fig. \ref{fig:localMin}a), the predicted takeover time is 0.97 s (ground truth = 0.89 s) and the variables in blue that push the takeover time low are “TBTC\&TBTB = 3.48”, “URG = 2”, “HAND = 0”, and “IRU = 0”  while the variable in red that pushes the predicted takeover time high is “AGE = 41”. 
In the second example, the predicted takeover time is 3.91 s (ground truth = 3.94 s) and the variables that push the takeover time high are “TOR\_V = 0”, “URG = 0”, “AGE = 40.9”, and “TBTC\&TBTB = 15” while the variables that push the takeover time low are “HAND = 0” and “SIM = 2” (see Fig. \ref{fig:localMin}b). The third example illustrates that the predicted takeover time is 7.88 s (ground truth = 8.70 s) and the variables that push the takeover time high are “TOR\_V = 1”, “SIM = 1”, “AGE = 38.5”, “URG = 0”, and “TBTC\&TBTB = 30” while the variable that pushes the predicted takeover time low is “HAND = 0” (see Fig. \ref{fig:localMin}c).

\section{DISCUSSION}
\subsection{Model Performance} 

In this work, we showed that XGBoost is an effective and efficient model to predict takeover time during the control transition period in conditionally automated driving. Our work used the mean takeover time reported in previous studies based on a meta-analysis study \cite{zhang2019determinants}. Collision risks during takeover are determined by individual takeover time rather than by the mean of the takeover time. However, the suggested methods in this work can be used as the baseline model, which can be personalized to predict individual takeover time based on the individual drivers’ characteristics and specific takeover scenarios.  
The performance of the XGBoost prediction model, including RMSE, Adj. $R^2$, MAE, and Corr is shown in Table \ref{table:modelcompperformance1}, using a 10-fold cross-validation strategy. We also performed a 10-fold cross-validation to compare the performance of XGBoost with other algorithms using Matlab Regression Learner (See Table \ref{table:modelcompperformance1}). XGBoost performed the best across all the metrics among the list of the models, including, linear regression, linear SVM, fine tree, and random forest. 
Also, we showed how the XGBoost model performed at different cumulative time bins (See Table \ref{table:modelcompperformance2}). Adj. $R^2$ was first increasing and then decreasing after it peaked at 0.625, while RMSE and MAE (min/max residual, which is defined as the min or max absolute difference between predicted takeover time and its ground truth) were increasing and gradually saturated when more samples with larger takeover time were added. This is possibly because there are increasingly fewer samples with larger takeover time (4 samples $>$ 6 s). Corr was consistently good across different cumulative time bins.

\begin{table}
\centering
\caption{Performance measures comparison between different models.}
\label{table:modelcompperformance1}
\resizebox{7cm}{!} {
\begin{tabular}{c c c c c}
\hline\hline
\small
\textbf{Models}      & \textbf{RMSE} $\downarrow$ & \textbf{Adj. $R^2$} $\uparrow$ & \textbf{MAE} $\downarrow$  & \textbf{Corr} $\uparrow$ \\ 
\hline
XGBoost              & \textbf{0.806} & \textbf{0.573}                     & \textbf{0.505} & \textbf{0.883}  \\
Linear regression  & 0.887          & 0.412                              & 0.672  & 0.642          \\
Linear SVM  & 0.917          & 0.372                              & 0.655 & 0.630         \\
Fine tree  & 1.017          & 0.321                              & 0.694     & 0.566     \\
Random forest  & 0.958          & 0.392                              & 0.677  & 0.648          \\
\hline\hline
\end{tabular}}
\end{table}

\begin{table}
\centering
\caption{Performance measures comparison of XGBoost at different cumulative time bins.}
\label{table:modelcompperformance2}
\resizebox{9cm}{!} {
\begin{tabular}{c c c c c c}
\hline\hline
\small
\textbf{Models (Samples)}      & \textbf{RMSE} $\downarrow$ & \textbf{Adj. $R^2$} $\uparrow$ & \textbf{MAE} $\downarrow$ & \textbf{Min/Max Residue} $\downarrow$ & \textbf{Corr} $\uparrow$ \\ 
\hline
XGBoost 0-2 s (151) & 0.267   & 0.393     & 0.198 & 0.185/0.212 & 0.828\\
XGBoost 0-3 s (368) & 0.360  & 0.531      & 0.274    & 0.263/0.292 & 0.869        \\
XGBoost 0-4 s (453) & 0.473          & 0.578         & 0.344 & 0.332/0.358 & 0.884  \\
XGBoost 0-5 s (486) & 0.556          & 0.605            &0.392 &0.376/0.410                  & 0.892          \\
XGBoost 0-6 s (508) & 0.639          & 0.625                              & 0.436 & 0.412/0.460  & 0.899         \\
XGBoost 0-7 s (515) & 0.739    & 0.583      & 0.476     &0.460/0.502                         & 0.887          \\
XGBoost 0-8 s (517) & 0.804    & 0.538      & 0.502    & 0.485/0.529                         & 0.874  \\
XGBoost 0-9 s (519) & 0.806          & 0.573 & 0.505                              & 0.490/0.534      & 0.883   \\
\hline\hline
\end{tabular}}
\end{table}

\subsection{Model Explainability}
We used SHAP explainer to understand the effect of the variables on the takeover time by providing global and local explanations. The provided global explanation helped us rank the variables based on their importance in the prediction of the takeover time. The importance of the predictor variables gave us direct guide in variable selection to improve the performance of XGBoost. We further explored the main and interaction effects of the most important variables on the takeover time prediction as evidenced in Figs. \ref{fig:Main Effect} and \ref{fig:Inter Effect}. For local explanation, SHAP provided prediction results by comparing the predicted value to a base value as a reference. Also, it showed the contributions of each variable-value set in pushing the results close to the ground truth as shown in Fig. \ref{fig:localMin}.
Furthermore, we found that the predictor variables selected were not exactly the same from \cite{zhang2019determinants}. In \cite{zhang2019determinants}, seven variables were included in their linear mixed model, i.e., LAD, TOR\_A, TOR\_VT, NDT\_V, HAND, URG, and IRU. Variables that did not have significant effects ($p>0.05$) on takeover time were excluded from the model. In our work, we used another variable selection method based on the importance of variables generated from SHAP. A sequential method by adding one variable at a time from the most important one was used to select the variables until RMSE stopped decreasing. The importance ranking was obtained by their prediction contributions based on SHAP values with local accuracy and consistency \cite{lundberg2017unified,lundberg2020local}, which had a solid theoretical support from game theory. The XGBoost model selected seven variables as shown in Table \ref{table:modelperformance}.

URG (Urgency) was shown to be the most important variable in the prediction of the takeover time. Similarly, Zhang et al. \cite{zhang2019determinants} found that URG had the highest correlation with the takeover time and a shorter takeover time was associated with a higher level of urgency of the takeover situation. In addition, we showed that URG interacted with TOR\_V (Visual TOR) (see Fig. \ref{fig:Inter Effect}a). At a high level of urgency, non-visual TORs increased the takeover time compared to those with visual TORs. This seemed to be counter-intuitive. We examined other variables when URG $=2$, TOR\_V $=1$ and we found that $44.1\%$ instances were also with driving simulators with a high fidelity while this percentage was $19.2\%$ when URG $=2$, TOR\_V $=0$. In Fig. \ref{fig:Inter Effect}f, driving simulators with high fidelity reduced the takeover time compared to those with low or medium fidelity. This might explain the results. At a low or medium level of urgency, visual TORs increased the takeover time compared to those with non-visual TORs. This is consistent with previous research that either auditory or vibrotactile TORs reduced takeover time compared to visual ones \cite{petermeijer2017take,roche2019should}.

The second most important variable was TBTC\&TBTB (Time budget to collision or to other boundaries). In agreement with previous studies, an increase in time budget led to an increase in takeover time \cite{gold2018modeling, mcdonald2019toward}. Also, TBTC\&TBTB were shown to interact with TOR\_V. At low time budget (i.e., high urgency), TOR\_V decreased the takeover time. The variable URG is negatively correlated with TBTC\&TBTB based on the coding protocol, i.e., it was considered as high urgency, when the time budget $\leq8$ s, medium urgency when it was between 8 s and 15 s, and low urgency when it was larger than 15 s. This, to a large extent, explained the interaction effect between TBTC\&TBTB and TOR\_V. 

The relationships between AGE and the takeover time seems complicated (Fig. \ref{fig:Main Effect}c), which might be one of the reasons that no consistent results were found in the literature by comparing old with young participants \cite{clark2017age,li2018investigation,korber2016influence}. We found that 45 was the turning point that for those younger than 45, age was positively correlated with the takeover time while for those older than 45, age was negatively correlated with takeover time. However, the majority of the participants were younger than 45 in the previous studies. More research should be conducted to understand the behavior of those older than 45 years old in the takeover transition period. Furthermore, we found that AGE interacted with TBTC\&TBTB (Fig. \ref{fig:Inter Effect}c) when it was smaller than 45. A smaller time budget decreased takeover time for those between 30 and 45 years old while a smaller time budget increased the takeover time for those younger than 30 years old. This result might be explained that compared to those under 30, those between 30 and 45 were more conservative and careful in automated driving compared to those below 30 \cite{tefft2012motor}. 

The next important variable was HAND (Hand holding a device). Consistent with previous studies (e.g., \cite{zhang2019determinants}), NDTs with a device in hands increased takeover time. In Fig. \ref{fig:Inter Effect}d, we showed that HAND interacted with TBTC\&TBTB. A larger time budget increased the takeover time when performing NDTs with a device in hands. This was in line with \cite{zhang2019transitions,wan2018effects,wandtner2018effects}. However, a large time budget decreased the takeover time compared to a small to medium time budget when the participants' hands were free during NDTs. This effect was not big by examining the SHAP values in Fig. \ref{fig:Inter Effect}d.

Our results showed that visual TORs increased takeover time, which was consistent with previous literature because visual alerts were less conspicuous compared to auditory and vibrotactile alerts \cite{petermeijer2017take}. The interaction effect between URG and TOR\_V was explained in Fig. \ref{fig:Inter Effect}a.

With regard to the simulator fidelity, a high fidelity reduced the takeover time compared to a low or medium fidelity, which might be explained by the difference of ecological validity in different types of simulators. In addition, we found that high urgency decreased the takeover time in low or medium fidelity simulators, while high urgency increased the takeover time in high fidelity simulators (see Fig. \ref{fig:Inter Effect}f). This effect seemed not reasonable for high fidelity simulators. One possible reason is that other factors potentially played a role and further studies are needed to investigate it. 

For IRU (Interaction with other road users), the model showed that the surrounding traffic increased the takeover time. This was consistent with previous literature which investigated the effects of traffic in the same and opposite direction on takeover time \cite{du2020evaluating,gold2016taking, radlmayr2014traffic}.
When there was more traffic in the surroundings, drivers needed more time to scan the environment, identify the potential hazards, and negotiate scenarios properly. No clear interaction effects with others were found.

\subsection{Implications}
 
To achieve reliable takeover time prediction, we recommend that the vehicle should have a systematic record of the driver's demographic information and the vehicle's settings, as well as interior camera/sensors to detect drivers' NDT status and exterior camera/sensors to determine environment complexity.

The results of our study provide the automotive industry with recommendations of the most important variables to predict takeover time. According to the predicted takeover time, adaptive in-vehicle alert systems can be developed to help drivers to intervene when necessary. For instance, when the model predicts that it would take drivers 8 seconds to take over control, advanced driver assistance systems should be initiated to help drivers identify the potential hazards in the environment and provide them with suggestions to shorten takeover time and improve takeover quality. 

Meanwhile, our method could be potentially employed to predict stabilisation time and specific driving behaviors following TORs \cite{melnicuk2021effect}, which provide a more comprehensive description of takeover performance. 
Broadly speaking, our findings can facilitate the interactions between drivers and automated vehicles, enhance driving safety in intelligent transportation systems, and improve automated vehicle acceptance across the whole population.

\subsection{Limitations}
First, although 18 variables were studied in this work, more variables in literature were identified that affected takeover time. For instance, drivers' emotional valence, cognitive load, and vehicle speed were found to influence takeover time and quality \cite{du2020evaluating,du2019examining}. Also, the vehicle operating state (e.g. traffic flow, acceleration) is another important factor affecting the takeover time, which can be included for prediction in the future studies. 

Second, the takeover time used in this study is based on the mean value. Collisions during takeovers are usually determined by individual takeover time and not by the mean of the takeover time. For instance, if the driver was sleeping during the ride, it might take him/her more time than usual to take over control. Thus, our study might not be suitable to identify the exact time span for a safe takeover. However, the proposed model can be used as the baseline and can be used to predict individual takeover time when personalized data is collected specific to the driver and the takeover scenarios in the future. Also, takeover performance is more than takeover time. Drivers' stabilisation time of both driver state and driving performance indicators, as well as different driving behaviors need to be considered to quantify takeover performance comprehensively. 

Third, the majority of the studies were conducted in a driving simulator which might not reflect the real situations and indicate a low external validity of the study. For instance, Carsten and Jamson \cite{carsten2011driving} discussed that in driving simulators people's perceived risk of the situation was lower compared to a real driving situation. 

Fourth, the interaction effects were only shown between two predictor variables in Fig. \ref{fig:Inter Effect} and those involving more than two variables were difficult to illustrate with SHAP. This might explain some of the counter-intuitive results shown previously.

\section{CONCLUSION}
In this work, we predicted takeover time using XGBoost and SHAP with good performance and explainability. Compared to previous studies that investigated the effects of individual factors on takeover time, our model is able to examine 18 factors at the same time and identified the most important factors (i.e., URG, TBTC\&TBTB, AGE, HAND, TOR\_V, SIM, AND IRU) in predicting takeover time with better performance. In addition, the model is able to show their main effects and interaction effects on takeover time.
Furthermore, our model is also able to provide explanations about individual prediction instances, which paves the way to design in-vehicle monitoring systems to improve takeover performance in conditionally automated driving.

\bibliography{main}



%


\begin{IEEEbiography}[{\includegraphics[width=1in,height=1.55in,clip,keepaspectratio]{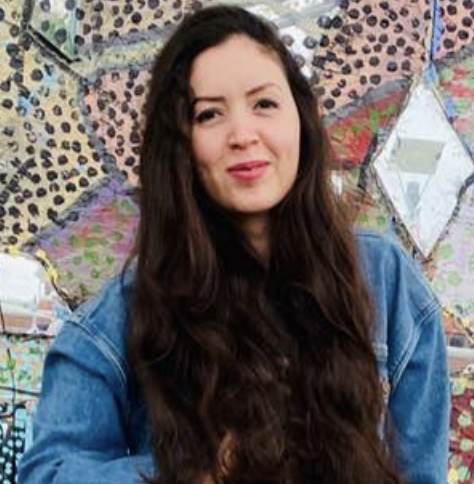}}]{Jackie Ayoub received her B.E. degree in mechanical engineering from Notre Dame University, Lebanon, in 2016 and her master degree in Industrial and Systems Engineering from University of Michigan, Dearborn, in 2017. She is currently pursuing her Ph.D. in Industrial and Systems Engineering in the University of Michigan, Dearborn. Her main research interests include human-computer interaction, human factors and ergonomics, and sentiment analysis.}
\end{IEEEbiography}
\vskip 0pt plus -1fil
\begin{IEEEbiography}[{\includegraphics[width=1in,height=1.55in,clip,keepaspectratio]{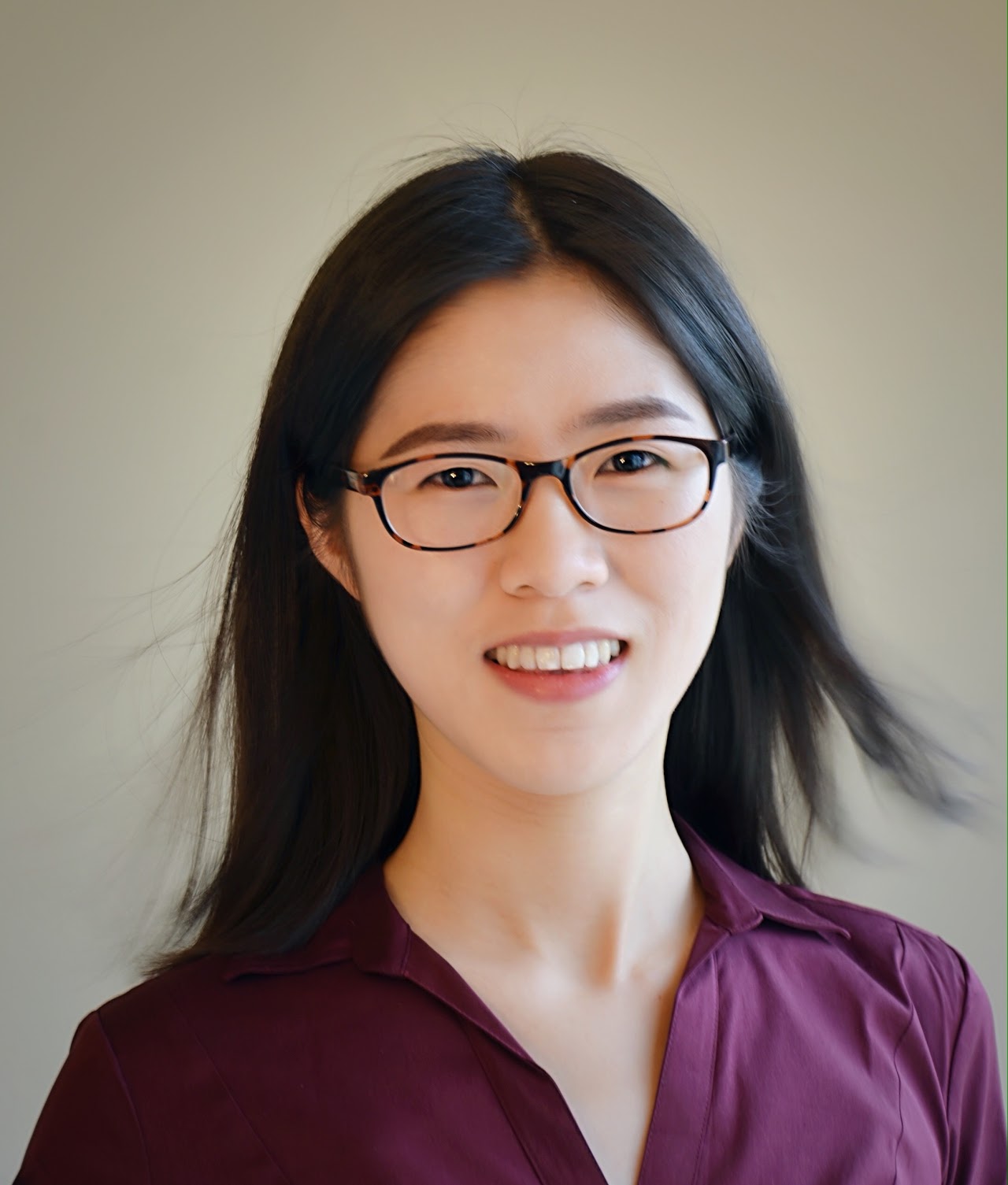}}]{Na Du is an Assistant Professor in the Department of Informatics and Networked Systems at the University of Pittsburgh. She received her Ph.D. degree in Industrial \& Operations Engineering from the University of Michigan and Bachelor’s degree in Psychology from Zhejiang University. Her research interests include human-computer interaction, transportation human factors, computational modeling of human behaviors, and human-centered design.}
\end{IEEEbiography}
\vskip 0pt plus -1fil
\begin{IEEEbiography}[{\includegraphics[width=1in,height=1.55in,clip,keepaspectratio]{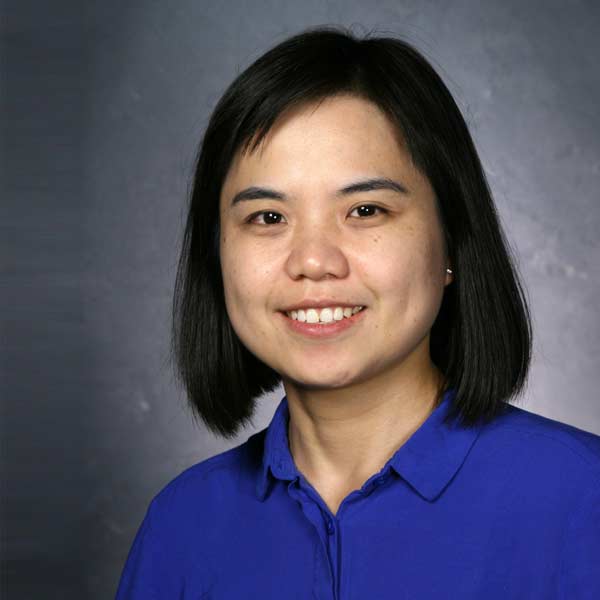}}]{X. Jessie Yang is an Assistant Professor in the Department of Industrial and Operations Engineering, University of Michigan, Ann Arbor. She earned a PhD in Mechanical and Aerospace Engineering (Human Factors) from Nanyang Technological University, Singapore. Dr. Yang’s research include human-autonomy interaction, human factors in high-risk industries and user experience design.}
\end{IEEEbiography}
\vskip 0pt plus -1fil
\begin{IEEEbiography}[{\includegraphics[width=1in,height=1.55in,clip,keepaspectratio]{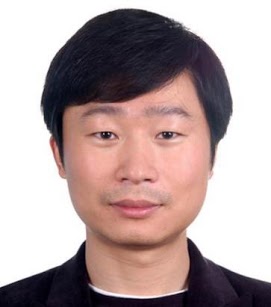}}]{Dr Feng Zhou received the Ph.D. degree in Human Factors Engineering from Nanyang Technological University, Singapore, in 2011 and Ph.D. degree in Mechanical Engineering from Gatech Tech in 2014. He was a Research Scientist at MediaScience, Austin TX, from 2015 to 2017. He is currently an Assistant Professor with the Department of Industrial and Manufacturing Systems Engineering, University of Michigan, Dearborn. His main research interests include human factors, human-computer interaction, engineering design, and human-centered design.}
\end{IEEEbiography}



\end{document}